\documentclass[sigconf]{acmart}
\usepackage{hyperref}
\usepackage{algorithm}
\usepackage{algpseudocode}
\usepackage{balance}
\usepackage{multirow}    
\usepackage{makecell}    
\usepackage{siunitx}     
\usepackage{wrapfig}    
\usepackage{listings}
\usepackage[table]{xcolor}
\usepackage{siunitx}
\definecolor{rowtiny}{RGB}{245,247,252}
\definecolor{rowsmall}{RGB}{232,236,248}
\definecolor{rowbase}{RGB}{215,222,240}
\usepackage{calc}
\newcommand{\AsymHL}[1]{%
  {\setlength{\fboxsep}{0pt}%
  \colorbox{blue!10}{%
    \hspace{-0.em}%
    \rule[-0.9ex]{0pt}{3.3ex}%
    \smash{#1}%
    \hspace{0.03em}%
  }%
  }%
}
\AtBeginDocument{%
  }

\copyrightyear{2026}
\acmYear{2026}
\setcopyright{cc}
\setcctype{by}
\acmConference[MM '26]{Proceedings of the 34th ACM International Conference on Multimedia}{November 10--14, 2026}{Rio de Janeiro, Brazil}
\acmBooktitle{Proceedings of the 34th ACM International Conference on Multimedia (MM '26), November 10--14, 2026, Rio de Janeiro, Brazil}
\acmDOI{10.1145/3767308.3835353}
\acmISBN{979-8-4007-2213-4/2026/11}

\begin{document}


\title{CrossMambaTuning: Synergistic Spatial and Cross-Layer Adaptation for Machine Vision Compression}

\author{Haobo Xiong}
\affiliation{%
  \institution{School of Computer Science and Technology, Xidian University}
  \city{Xi'an}
  \country{China}
}
\email{24031110055@stu.xidian.edu.cn}
\author{Shaobo Liu}
\affiliation{%
   \institution{School of Computer Science and Technology, Xidian University}
  \city{Xi'an}
  \country{China}
}
\email{shaoboo.liu@stu.xidian.edu.cn}

\author{Kai Liu}
\affiliation{%
   \institution{School of Computer Science and Technology, Xidian University}
  \city{Xi'an}
  \country{China}
}
\email{kailiu@mail.xidian.edu.cn}
\author{Chongyang Ding}
\authornote{Corresponding author.}
\affiliation{%
   \institution{School of Computer Science and Technology, Xidian University}
  \city{Xi'an}
  \country{China}
}
 \email{dingcy@xidian.edu.cn}
\renewcommand{\shortauthors}{Haobo Xiong, Shaobo Liu, Kai Liu, and Chongyang Ding}


\begin{abstract}
To reduce deployment cost and retraining overhead, adapting pretrained learned image compression (LIC) models to downstream machine vision tasks has attracted growing attention. However, existing methods typically insert fine-tuning modules independently into frozen backbones, lacking explicit mechanisms for cross-layer coordination. To address this limitation, we propose a novel framework named CrossMambaTuning, which integrates State Space Models with cross-layer interaction mechanisms for parameter-efficient fine-tuning. Specifically, we design an efficient Mamba adapter equipped with task-specific prompts and multi-scale branching to precisely capture both local features and global dependencies. Furthermore, we introduce a Scale-Invariant Cross-Layer Adapter (SICA) utilizing a parameter-sharing strategy to fuse task information across different scales and reduce redundancy. Extensive experiments demonstrate that CrossMambaTuning achieves state-of-the-art (SOTA) performance on multiple machine vision tasks, reducing parameter overhead by 72\% compared to SOTA methods. \textbf{Code is available at \url{https://github.com/rsr1123/CrossMambaTuning}.}
\end{abstract}


\maketitle


\section{Introduction}
\begin{figure}[!ht]
  \centering

    \centerline{\includegraphics[width=0.95\linewidth]{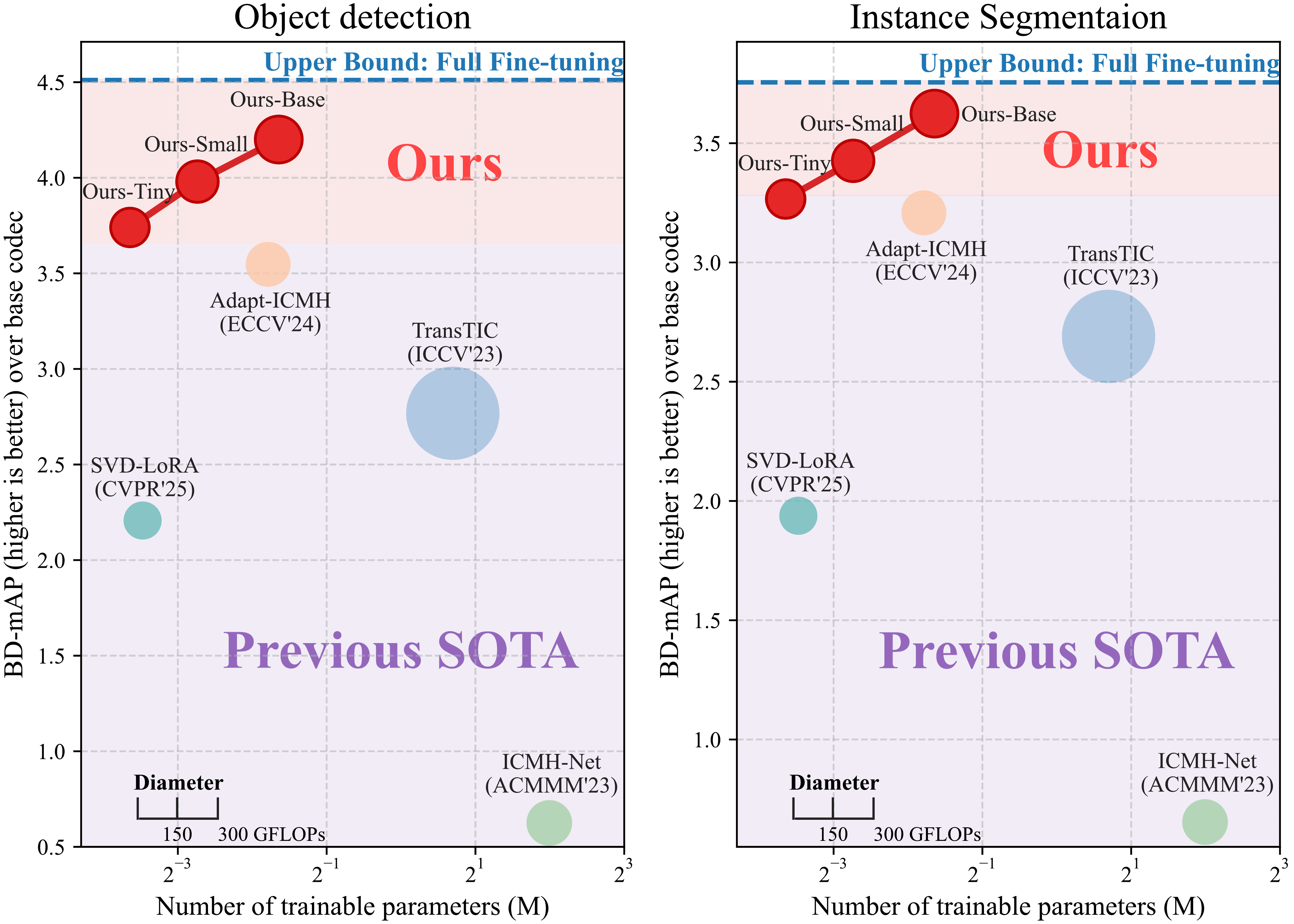}}
    \caption{
      Performance comparison on COCO2017-val \cite{lin2014microsoft}. The different variants of the proposed method, full fine-tuning, TransTIC \cite{chen2023transtic}, ICMH-Net \cite{liu2023icmh},Adapt-ICMH \cite{li2024image}, and SVD-LoRA \cite{LoRA_comp}, are compared. BD-mAP is computed using the base codec of TIC \cite{lu2021transformer} as the anchor. The size of the circles represents Gflops for encoding.
    }
    \label{mAP_paras}
\end{figure}

﻿
The exponential growth of visual data used in downstream applications \cite{DBLP:conf/cvpr/HeZRS16,DBLP:conf/nips/RenHGS15,DBLP:conf/nips/XieWYAAL21}, such as the Internet of Things (IoT) \cite{DBLP:conf/isicir/Chuah14}, demands efficient compression techniques that support machine vision tasks. However, practical deployment scenarios often require compression models to preserve image reconstruction capability for human viewing. To minimize the parameter overhead and retraining cost of maintaining multiple task-specific systems, early studies \cite{choi2022scalable,DBLP:conf/vcip/LiuJFCZ23} explored unified frameworks. However, these approaches typically required training multi-task networks from scratch, which incurred substantial training and storage overheads. Therefore, reducing the training and storage overhead for diverse applications has become an important direction.

To address this need, recent studies \cite{chen2023transtic,liu2023icmh,li2024image} increasingly adopt the parameter-efficient fine-tuning (PEFT) paradigm, where lightweight task-specific modules are attached to pretrained learned image compression (LIC) models. Instead of training separate task-specific codecs from scratch, these methods reuse pretrained models and introduce only minimal additional parameters for downstream machine vision adaptation. In practical deployment, adapting a pretrained LIC model to machine vision tasks avoids storing an additional full set of model weights, thereby substantially reducing both retraining costs and storage overhead.

Building on the PEFT paradigm, existing methods can be broadly categorized into two approaches. The first category (e.g., ICMH-Net \cite{liu2023icmh} and TransTIC \cite{chen2023transtic}) employs masks or visual prompting\cite{bar2022visual}. However, these methods often impose rigid structural constraints on the encoder (e.g., TransTIC is exclusive to Transformers), limiting their architectural flexibility. The second category focuses on architecture-agnostic fine-tuning via Low-Rank Adaptation (LoRA) \cite{hu2022lora,LoRA_comp} or adapters \cite{HoulsbyGJMLGAG19,li2024image}. 
Notably, methods like SFMA achieve efficient adaptation via input-adaptive frequency modulation, but frequency modulation often fails to capture long-range spatial dependencies. 
Although attention mechanisms \cite{DBLP:journals/corr/BahdanauCB14,DBLP:conf/nips/VaswaniSPUJGKP17} could address this limitation, their quadratic complexity makes them intractable for high-resolution images. 
A promising alternative lies in State Space Models \cite{DBLP:journals/corr/abs-2312-00752,ZhuL0W0W24,DBLP:conf/nips/LiuTZYX0YJ024}, such as VMamba \cite{DBLP:conf/nips/LiuTZYX0YJ024}, which capture long-range dependencies with linear complexity, offering an efficient solution for global spatial modeling.

Furthermore, existing methods \cite{li2024image,LoRA_comp} typically insert modules independently at various layers within the network, without explicit mechanisms for cross-layer coordination. Consequently, task-relevant information may not be effectively integrated across layers, limiting the utilization of complementary semantics.

To address these challenges, we propose CrossMambaTuning, a parameter-efficient adaptation framework for machine vision tasks. First, we design an efficient Mamba adapter equipped with task-specific prompts. Simultaneously, this module utilizes multi-scale branching to extract local features and selective scanning to capture long-range spatial dependencies, enabling precise modeling of task-relevant priors. 
Second, to alleviate the layer-isolation issue, we introduce a Scale-Invariant Cross-Layer Adapter (SICA) with a parameter-sharing strategy.
This approach implements a scale-agnostic cross-layer mechanism to fuse task information across features of different scales. 
Such a design facilitates cross-layer information fusion and reduces compression redundancy, thereby significantly improving parameter utilization efficiency.

Leveraging these innovations, our framework achieves parameter efficiency. The Tiny variant reduces parameter overhead by 72\% compared to the state-of-the-art (SOTA) method \cite{li2024image}, using 0.08M parameters while achieving comparable or superior performance.

In summary, the main contributions are as follows:
\begin{itemize}
\item We propose an efficient Mamba adapter augmented with task-specific prompts and local multi-scale branching modules. By effectively modeling spatial dependencies within images, this design facilitates efficient adaptation of LIC models to downstream machine vision tasks.
\item We introduce a cross-layer information fusion mechanism employing a parameter-sharing strategy. By implementing a Scale-Invariant Cross-Layer Adapter, this mechanism effectively fuses task information across different scales, thereby reducing redundancy and enhancing parameter efficiency.
\item We present the CrossMambaTuning framework by integrating the proposed Mamba adapter and cross-layer fusion mechanism. Extensive experiments demonstrate that our method achieves SOTA performance on multiple machine vision tasks across various network configurations.
\end{itemize}



\section{Related Work}

\subsection{Learned Image Compression (LIC)}

Learned Image Compression (LIC) \cite{balleEndendOptimizedImage2017a,balleVariationalImageCompression2018a,minnenJointAutoregressiveHierarchical2018,heCheckerboardContextModel2021,jiang2023mlicpp} demonstrates strong competitiveness against traditional codecs such as VVC \cite{bross2021overview} and HEVC \cite{sullivan2012overview}, typically evaluated using PSNR and MS-SSIM. LIC models generally consist of a transform module and an entropy model. Prior work on transform modules focuses on nonlinear transformation, adopting convolutions with GDN \cite{balleVariationalImageCompression2018a} or enhancing expressiveness via residual learning and attention \cite{chengDeepResidualLearning2019,zou2022devil}. Recent studies use Transformer-based architectures to model long-range dependencies \cite{lu2021transformer,liu2021swin}. For entropy models, hyperprior-based methods capture spatial dependencies \cite{balleVariationalImageCompression2018a}, while autoregressive models exploit decoded context \cite{minnenJointAutoregressiveHierarchical2018}. Recent work integrates richer cues for entropy estimation \cite{jiang2023mlicpp}.

\subsection{Machine Vision in LIC}
Despite the success of LIC models optimized for human perception, their compressed representations are not always well aligned with the requirements of downstream machine vision tasks. Early studies \cite{bai2022towards,choi2022scalable} explored unified or multi-task frameworks to support diverse usage scenarios within a single model.
For instance, Choi \& Bajic \cite{choi2022scalable} proposed a scalable coding framework via latent decomposition, while Zhang et al.\cite{zhang2024allinone} presented an ensemble framework with multi-path aggregation. However, these multi-branch designs incur substantial training and storage overheads, limiting practical deployment. 
Consequently, recent research \cite{chen2023transtic,li2024image,LoRA_comp} has shifted toward the fine-tuning paradigm to adapt pretrained LIC models. These approaches typically fall into two categories. The first utilizes masks or visual prompts, such as the mask generator \cite{fischer2022boosting} and TransTIC \cite{chen2023transtic}. However, such methods are often tied to specific architectures (e.g., TransTIC is restricted to Transformer-based models), limiting their versatility. The second category employs Low-Rank Adaptation (LoRA) \cite{LoRA_comp} or Adapters \cite{li2024image}. By inserting lightweight task-specific modules into frozen codecs, these methods apply to a broad range of backbones, achieving strong task performance with minimal additional parameters.
\subsection{Mamba-based PEFT Methods}
Recently, State Space Models (SSMs) \cite{ZhuL0W0W24,DBLP:conf/nips/LiuTZYX0YJ024,DBLP:conf/cvpr/Xie0TZZ25,DBLP:conf/iclr/YoshimuraHM25} have shown strong potential in vision tasks, offering global modeling capacity competitive with Transformers while maintaining linear computational complexity. Leveraging their linear complexity and efficient long-range modeling capability, recent studies have explored the use of SSMs as lightweight adaptation modules in pretrained models. In medical image segmentation, Triplane Mamba \cite{DBLP:conf/miccai/WangLDL24} introduces Mamba as an adapter for 3D SAM \cite{DBLP:journals/corr/abs-2304-13785}, enabling parameter-efficient fine-tuning through long-range spatial dependency modeling in 3D data. In point cloud understanding, PMA \cite{DBLP:conf/cvpr/ZhaWG000OYCX25} proposes a Point Mamba Adapter that constructs an ordered feature sequence from all layers of the pretrained model and uses Mamba for cross-layer semantic fusion. Mamba has also been adopted for cross-model adaptation. For example, MAVLT \cite{DBLP:journals/tcsv/ShiZLHMS25} uses a Mamba adapter as a bridge between the visual and language encoders, achieving efficient vision-language fusion with only a small number of trainable parameters.  These studies suggest that Mamba can serve not only as a backbone but also as an adapter in pretrained models.

Unlike existing Mamba-based adaptation methods that are largely tailored to specific scenarios, our approach is developed for machine vision compression through parameter-efficient adaptation of pretrained LIC models. It provides a unified framework for machine vision compression across classification, object detection, and instance segmentation, and is applicable to diverse architectures, including both CNN-based and Transformer-based models. Its strong performance comes from two complementary components: the Task-aware Mamba Adapter performs efficient spatial modeling, while SICA alleviates the  redundancy introduced by frozen codecs. 
Notably, the Tiny variant requires only 0.08M parameters while maintaining competitive performance across multiple tasks.
﻿

%
\section{Method}

\subsection{Overview and Motivations}

\begin{algorithm}[tb]
   \caption{Procedure of CrossMambaTuning}
   \label{alg:cross_mamba_tuning}
\begin{algorithmic}[1]
   \Require Image $x$, Frozen Backbone $\{\mathcal{F}_k\}_{k=1}^{K}$, Mamba Adapters $\{\mathcal{A}_k\}_{k=1}^{K-1}$, SICA Module $\mathcal{T}(\cdot; \Theta)$ with gates $\{\alpha_k\}$, Global Code $E_{task}$, Layer Embeddings $E_{\text{layer}}^{(l)}$, Prompt Generator $\mathcal{G}$
   \Ensure Representation $y$
   \State $h \leftarrow \mathcal{F}_1(x)$
   \State $ P^{(1)} \leftarrow \mathcal{G}(E_{task} + E_{\text{layer}}^{(1)})$
   \State $z_1 \leftarrow \mathcal{A}_1(h; \text{prefix}= P^{(1)})$
   
   \For{$k = 1, \dots, K-2$}
      \State $h_{\text{frozen}} \leftarrow \mathcal{F}_{k+1}(z_k)$
      
{\setlength{\fboxsep}{0pt}
\State \colorbox{blue!10}{\hspace{-0.em}\rule[-0.8ex]{0pt}{2.5ex}\smash{$h_{\text{bypass}} \gets \alpha_k \cdot \mathcal{T}(z_k;\Theta)$}\hspace{0.03em}}
}
      \State $z_{in} \leftarrow h_{\text{frozen}} + h_{\text{bypass}}$
      
\State \AsymHL{$P^{(k+1)} \leftarrow \mathcal{G}(E_{task} + E_{\text{layer}}^{(k+1)})$}
      \State $z_{k+1} \leftarrow \mathcal{A}_{k+1}(z_{in}; \text{prefix}= P^{(k+1)})$
   \EndFor
   \State return $y \leftarrow \mathcal{F}_{K}(z_{K-1})$
\end{algorithmic}
\end{algorithm}

Existing PEFT-based adaptation frameworks exhibit two key limitations when adapting pretrained learned image compression (LIC) models to downstream machine vision tasks. First, current methods often struggle to capture long-range spatial dependencies, with underrepresented correlations between distant regions hindering the preservation of task-relevant global structure. Second, the lack of explicit cross-layer coordination may cause adapters at different layers to learn overlapping feature patterns \cite{DBLP:conf/naacl/LiZ25}, introducing redundancy and limiting the coordinated utilization of complementary semantics across layers.
To address these issues, we propose CrossMambaTuning, a PEFT framework tailored for machine vision. The framework comprises two lightweight modules that enhance spatial dependency modeling and cross-layer information integration:

\textbf{(i) Task-Aware Mamba Adapter.} We insert a Mamba-based adapter after each encoder/decoder stage except the last. Leveraging the SSM, the adapter captures long-range spatial dependencies efficiently. To incorporate task awareness, we employ a shared \textbf{Task-Specific Prompt Generator (TSPG)} to inject task priors and promote consistent information sharing across adapters.

\textbf{(ii) Scale-Invariant Cross-Layer Adapter (SICA).} To facilitate cross-layer interaction, we establish information pathways between adjacent layers. By aligning features across scales and extracting task-relevant semantics, this adapter enables cross-scale feature integration and facilitates information exchange across layers.

\begin{figure}[!t]
\centering
\centerline{\includegraphics[width=0.95\linewidth]{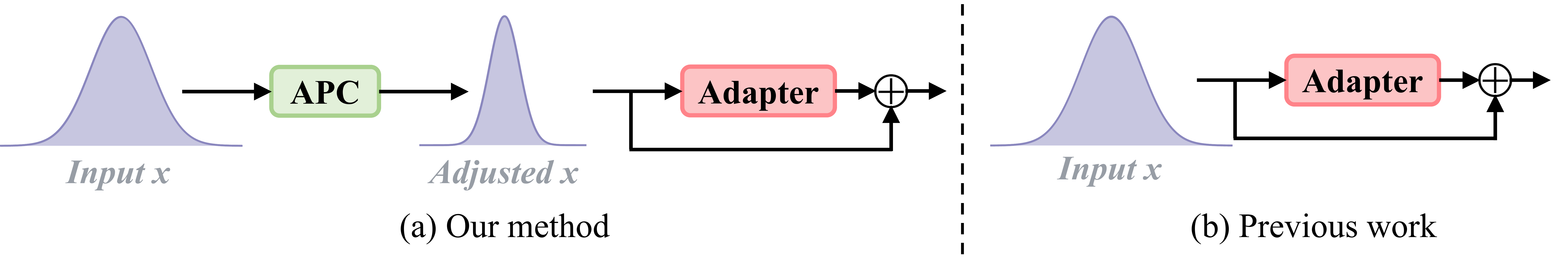}}
\caption{The APC modifies the input $x$ before the adapter. 
}
\label{APC}
\end{figure}
\begin{figure*}[!t]
  \centering
    \centerline{\includegraphics[width=0.9\linewidth]{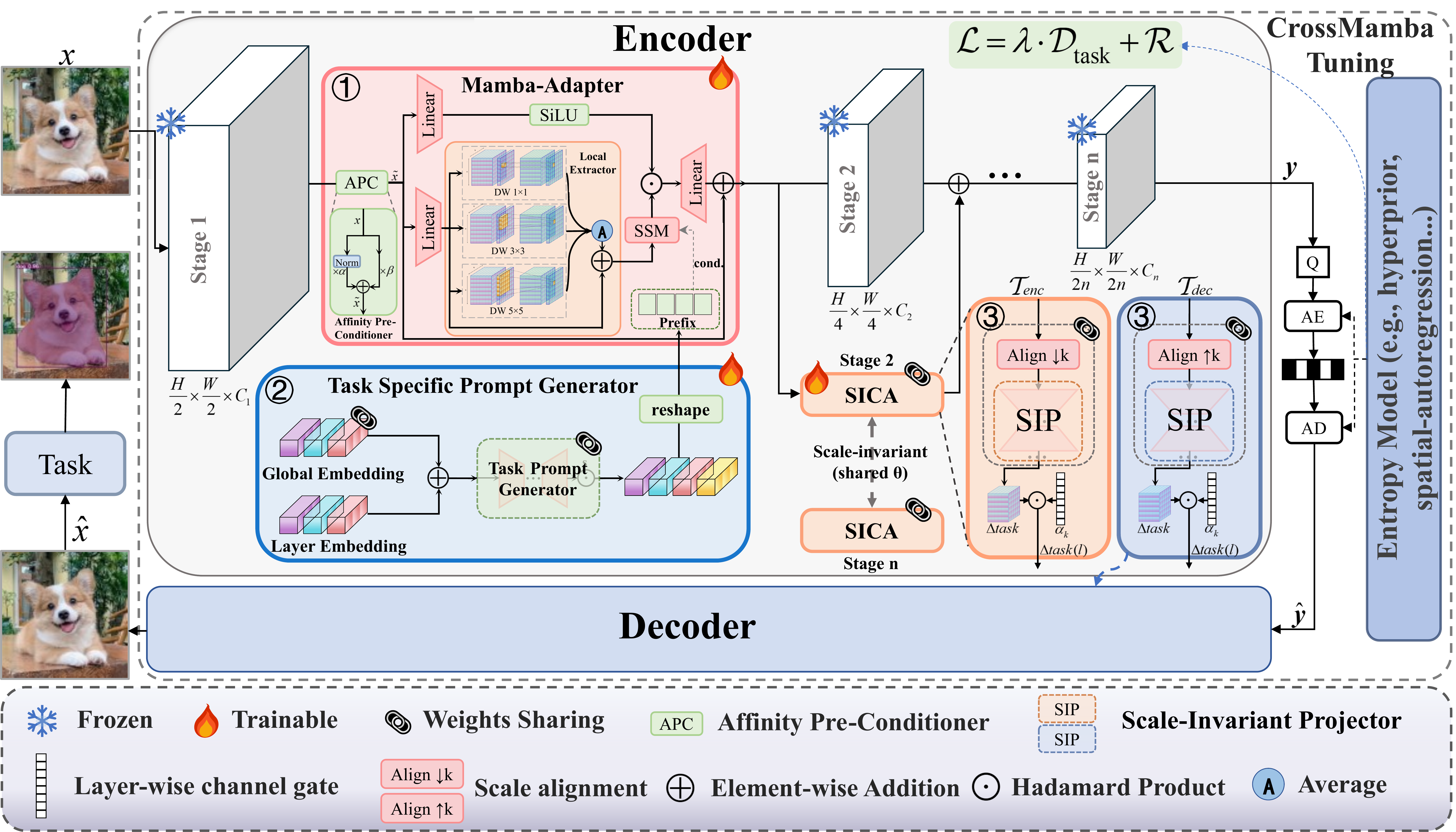}}
    \caption{
      Overview of the proposed CrossMambaTuning.The snowflake symbol represents frozen layers, while the flame symbol represents trainable layers. Since the decoding stage is the inverse process of the encoding stage, only the specific fine-tuning modules within the encoder are shown. More details can be found in Appendix H.
    }
    \label{main_process}
\end{figure*}

The procedure of the adaptation can be seen in Algorithm \ref{alg:cross_mamba_tuning}, while 
Figure \ref{main_process} provides an overview of the pipeline. Given an input image $x$, the encoder produces a compact latent $y$, which is coded into a binary bitstream via an entropy model; the decoder then reconstructs the image $\hat{x}$ optimized for the downstream task. Overall, CrossMambaTuning improves adaptation and compression efficiency while introducing only 1\%–4\% additional parameters.

\subsection{Task-aware Mamba Adapter}

To effectively capture long-range spatial dependencies while promoting interaction among adapters, we integrate the proposed Task-aware Mamba Adapter into the pre-trained codecs. As shown in Figure~\ref{main_process}, the adapter first applies an Affinity Pre-conditioner (APC) to the input features $x$ to align them with the distribution of downstream tasks. The resulting features $\tilde{x}$ are then projected into a lower-dimensional space through a linear transformation. 
The transformed features are processed by two parallel branches. One branch uses a local information extractor to capture multi-scale spatial details, followed by an SSM augmented with task-specific prompts to generate modulation signals, which are applied to the other branch via element-wise gating. An up-projection layer restores the original feature dimension, and the output is combined with the APC-adjusted features through a residual connection to produce the final representation.
Specifically, we employ three distinct designs to construct an efficient Mamba adapter:

\textbf{(i) Affinity Pre-conditioner (APC).}
Conventional PEFT methods typically apply additive modulation to intermediate features. However, the feature distribution generated by the frozen backbone often deviates from the optimal distribution for downstream tasks. The absence of proper preprocessing consequently limits the fine-tuned distribution from reaching the ideal state. To rectify this misalignment and thereby enhance the affinity for downstream tasks, we introduce the Affinity Pre-conditioner (APC).

﻿
As shown in Figure \ref{APC}, APC operates as a lightweight feature rectification module preceding the residual adapter. APC adopts a dual-path mechanism fusing the raw input with normalized features. The pre-conditioned feature $x_{\mathrm{APC}}$ is formulated as:
\begin{equation}
x_{\mathrm{APC}} = \text{Norm}(x) \odot \alpha + x \odot \beta,
\end{equation}
where $\text{Norm}(\cdot)$ denotes normalization such as LayerNorm \cite{DBLP:journals/corr/BaKH16}, and $\alpha, \beta$ are learnable scaling parameters.
The APC module introduces only $4C$ learnable parameters, where $C$ denotes the channel dimension; for Lu2022-TIC \cite{lu2021transformer} with $C=128$, this corresponds to 512 parameters per module, resulting in negligible parameter overhead.

﻿

As shown in Table \ref{tab:alignment_metrics}. We calculate the Wasserstein Distance and KL Divergence between the feature distributions of the full fine-tuning model and the Mamba-Adapter, with and without APC. We report the metrics for the latent representation $y$ and the average across network stages. The results indicate that APC mitigates distribution shift and aligns features closer to the full fine-tuning.

\begin{table}[h]
\centering
\caption{We measure the distribution distance relative to the full fine tuning (lower is better). The metrics are reported for the average of intermediate stages (Stages) and the latent ($y$).}
\label{tab:alignment_metrics}
\begin{tabular}{lcccc}
\toprule
\multirow{2}{*}{\textbf{Method}} & \multicolumn{2}{c}{\textbf{Wasserstein} $\downarrow$} & \multicolumn{2}{c}{\textbf{KL Divergence} $\downarrow$} \\
& Stages & $y$ & Stages & $y$ \\
\cmidrule(lr){1-5}
w/o APC & 0.0308 & 0.0388 & 0.0396 & 0.0163 \\
\rowcolor{blue!10} w/ APC & \textbf{0.0277} & \textbf{0.0318} & \textbf{0.0318} & \textbf{0.0077} \\
\bottomrule
\end{tabular}
\end{table}

\textbf{(ii) Local Information Extractor (LIE).} 
To mitigate the limited ability of SSMs \cite{DBLP:journals/corr/abs-2312-00752,DBLP:conf/nips/LiuTZYX0YJ024,ZhuL0W0W24,DBLP:conf/aaai/Pei0X25} to capture fine-grained local details, we introduce a Local Information Extractor to replace the original $3 \times 3$ depth-wise convolution. The module adopts a multi-branch design with depth-wise convolution kernels of sizes $1\times1$, $3\times3$, and $5\times5$, capturing pixel-level variations, local neighborhood context, and broader texture and edge patterns, respectively. Through multi-scale fusion, the extractor provides a strong local inductive bias for the subsequent SSM, enhancing its capacity to process high-frequency information such as edges and textures.

Formally, given an input $x \in \mathbb{R}^{B \times C \times H \times W}$, three depth-wise convolution branches independently process $x$ with kernel sizes $k \in {1, 3, 5}$. 
The refined feature $y$ is obtained by integrating multi-scale local information into the input:
\begin{equation}
y = x + \frac{1}{3} \sum_{k \in {1,3,5}} \text{DWConv}{k \times k}(x),
\end{equation}
where $\text{DWConv}{k \times k}(\cdot)$ denotes a depth-wise convolution with kernel size $k \times k$. 

\textbf{(iii) Task-Specific Prompt Generator (TSPG).} To condition the Mamba Adapters with task-oriented prior, we introduce the Task-Specific Prompt Generator (TSPG). 

Unlike instance-wise conditioning, TSPG generates a task prior that initializes the SSM states. Formally, we create a global shared task embedding $E_{\text{task}} \in \mathbb{R}^d$ to unify task objectives, and layer-specific embeddings $\{E_{\text{layer}}^{(l)}\}_{l=1}^L \in \mathbb{R}^d$ to capture layer-wise variations. The prompt $P^{(l)}$ for the $l$-th layer is generated via a lightweight projector $\mathcal{G}(\cdot)$:
\begin{equation}P^{(l)} = \mathcal{G}(E_{\text{task}} + E_{\text{layer}}^{(l)}).\end{equation}
This prompt is prepended to the input sequence $X^{(l)}$:
\begin{equation}X_{\text{in}}^{(l)} = [P^{(l)}; X^{(l)}],\end{equation}
where $[\cdot;\cdot]$ denotes concatenation along the sequence dimension. Consequently, the Mamba Adapter processes the prompt first, thereby initializing its hidden states with task-specific information before handling image features. Furthermore, the shared $E_{\text{task}}$ implicitly links distributed adapters, enabling effective joint optimization.

\subsection{Scale-Invariant Cross-Layer Adapter (SICA)}

\subsubsection{Problem Formulation}
We denote the frozen backbone as $\Phi$, comprised of $K$ consecutive stages. At the $k$-th stage, the feature extraction is governed by the frozen transformation operator $\mathcal{F}_k: \mathbb{R}^{C_{k-1} \times H_{k-1} \times W_{k-1}} \to \mathbb{R}^{C_k \times H_k \times W_k}$. The feature transition is formalized as:
\begin{equation}
z_k = \mathcal{F}_k(z_{k-1}), \quad \text{where } z_k \in \mathbb{R}^{C_k \times H_k \times W_k}.
\end{equation}
In conventional frameworks \cite{li2024image,LoRA_comp}, the adapted feature $z_k^{\text{adapt}} \in \mathbb{R}^{C_k \times H_k \times W_k}$ is introduced locally in a layer-wise manner and subsequently processed by $\mathcal{F}_k$. 
 Such a design lacks an explicit mechanism for cross-layer coordination and could limit the coordinated utilization of task-relevant information across layers. Moreover, adapters at different layers may learn overlapping feature patterns \cite{DBLP:conf/naacl/LiZ25}, thereby introducing redundancy.

\subsubsection{Design of the SICA Bypass Operator}
To address this, SICA establishes an explicit bypass connection. Let $z_k^{\text{adapt}}$ denote the adapted feature at stage $k$. The input to the next stage, denoted as $z_{k+1}^{\text{in}} \in \mathbb{R}^{C_k \times H_{k+1} \times W_{k+1}}$, is given by:
\begin{equation}
z_{k+1}^{\text{in}} = \underbrace{\mathcal{F}_k(z_k^{\text{adapt}})}_{\text{Frozen Path}} + \underbrace{\alpha_k \cdot \mathcal{T}(z_k^{\text{adapt}}; \Theta)}_{\text{SICA Bypass}},
\end{equation}
where $\alpha_k \in \mathbb{R}^{1\times C_k}$ is a learnable scalar initialized to 0.

To achieve extreme parameter efficiency, we employ a \textit{Scale-Invariant Parameter Sharing} strategy. We define the transformation operator $\mathcal{T}: \mathbb{R}^{C_k \times H_k \times W_k} \to \mathbb{R}^{C_k \times H_{k+1} \times W_{k+1}}$ as a composition of spatial and semantic operations:
\begin{equation}
\mathcal{T}(z) \triangleq (\mathcal{P}_{\text{inv}} \circ \mathcal{S}_{\text{align}})(z).
\end{equation}
The parameter set $\Theta$ is shared across stages. Specifically, we define $\Theta = \{\Theta_{\text{enc}}, \Theta_{\text{dec}}\}$, where $\Theta_{\text{enc}}$ applies to encoder and $\Theta_{\text{dec}}$ to decoder. The operators are defined as follows:

     \textbf{Scale-Alignment Spatial Operator} $\mathcal{S}_{\text{align}}$. To align resolutions, $\mathcal{S}_{\text{align}}$ applies distinct operations for encoder and decoder stages ($m \in \{\text{enc}, \text{dec}\}$) :
    \begin{equation}
    \mathcal{S}_{\text{align}}(z) = 
    \begin{cases} 
    \text{DWConv}_{s=2}(z), & \text{if } m = \text{enc} \\
    \text{DWConv}_{s=1}(\text{Up}(z)), & \text{if } m = \text{dec}
    \end{cases}
    \end{equation}
    where $\text{DWConv}_{s}$ denotes a depthwise convolution with stride $s$, and $\text{Up}(\cdot)$ represents $2\times$ bilinear interpolation. 

 \textbf{Scale-Invariant Projector} $\mathcal{P}_{\text{inv}}$. After spatial alignment, the features are processed via a bottleneck MLP. With reduction ratio $r$, the projection matrices are $\mathbf{W}_{\text{down}} \in \mathbb{R}^{\frac{C}{r} \times C}$ and $\mathbf{W}_{\text{up}} \in \mathbb{R}^{C \times \frac{C}{r}}$. The operator acts on the feature $z$ as:
\begin{equation}
\mathcal{P}_{\text{inv}}(z) = \mathbf{W}_{\text{up}} (\sigma \left( \mathbf{W}_{\text{down}} (z) \right)),
\end{equation}
where $\sigma(\cdot)$ denotes the SiLU activation \cite{elfwing2018sigmoid} . By sharing ${\mathbf{W}_{\text{down}}, \mathbf{W}_{\text{up}}}$ across all stages, $\mathcal{P}_{\text{inv}}$ captures semantic representations independent of spatial scale.

The same $\Theta$ generalizes across domains $\Omega_k \subset \mathbb{R}^{H_k \times W_k}$ of varying scales, enforcing a universal feature injection mechanism.
\subsection{Training Loss}
During the training phase, the weights of the pretrained codec are frozen, and only the inserted adapters are optimized. The overall training objective follows a RD optimization, defined as:
\begin{equation}
\mathcal{L} = \mathcal{R} + \lambda \cdot \mathcal{D}_{\text{task}},
\end{equation}
where $\mathcal{R}$ denotes the bitrate estimated by the entropy model (for the hyperprior entropy model, $\mathcal{R} = \mathcal{R}(\hat{\mathbf{y}}) + \mathcal{R}(\hat{\mathbf{z}})$), $\mathcal{D}_{\text{task}}$ represents the task specific perceptual distortion metric (defined in Appendix D), and $\lambda$ is the hyperparameter controlling the rate distortion tradeoff.

%
\begin{figure*}[!ht]
	\centering
	\includegraphics[width=1\linewidth]{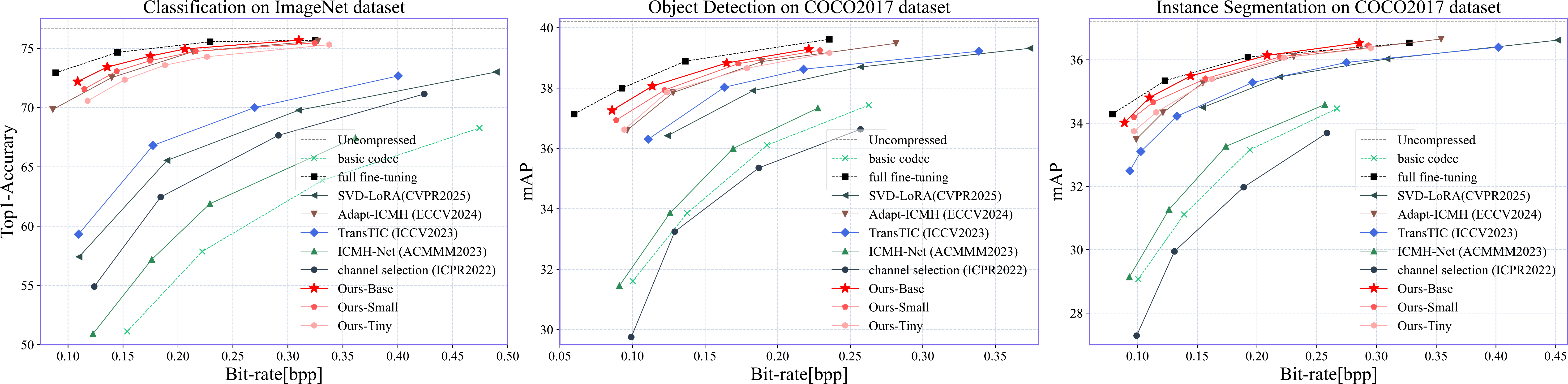}
	\caption{Rate-Accuracy performance comparison for different machine vision tasks using the Lu2022-TIC base codec.}
	\label{fig:main_tic}
	
\end{figure*}

\begin{table*}[!ht]
	\centering
	\caption{Performance comparison of different methods on three machine tasks, using TIC as the base codec. We report the number of trainable parameters and two BD metrics \citep{bjontegaard2001calculation}: BD-rate and BD-acc/mAP. The best and second-best results are highlighted in bold and underline, respectively.}
	\begin{tabular}{%
			c
			c
			cccccc
			r
		}
\toprule
\multirow{2}{*}{\textbf{Method}} & \multirow{2}{*}{\textbf{Venue}}
& \multicolumn{2}{c}{\textbf{Image Classification}}
& \multicolumn{2}{c}{\textbf{Object Detection}}
& \multicolumn{2}{c}{\textbf{Instance Segmentation}}
& \multirow{2}{*}{\makecell{\textbf{Trainable}\\\textbf{Params} $\downarrow$ (M)}} \\
\cmidrule(lr){3-4}\cmidrule(lr){5-6}\cmidrule(lr){7-8}
&& BD-rate$\downarrow$ & BD-acc$\uparrow$
& BD-rate$\downarrow$ & BD-mAP$\uparrow$
& BD-rate$\downarrow$ & BD-mAP$\uparrow$ \\
\midrule
		full fine-tuning&--
		&    /\   & 17.688 & -73.943\% & 4.511 & -67.977\% & 3.755& 7.51(100.00\%) \\
		\cmidrule(lr){1-9}
		channel selection \cite{liu2022improving}&ICPR'22
		& -37.178\% & \phantom{-}6.278  & \phantom{---}6.849\%   & -0.550\phantom{-} & \phantom{-}16.511\% & -0.949\phantom{-}& 0.92(12.25\%) \\
		ICMH-Net \cite{liu2023icmh}&ACM MM'23
		& -18.759\% & \phantom{-}3.360  & \phantom{-}-9.080\%  & 0.625 & -10.772\% & 0.654& 3.98(53.00\%) \\
		TransTIC \cite{chen2023transtic}&ICCV'23
		& -58.529\% & \phantom{-}9.956  & -46.301\% & 2.768 & -46.521\% & 2.690 & 1.62(21.57\%) \\
		Adapt-ICMH \cite{li2024image}&ECCV'24
				& -88.573\% & 16.901 & -55.150\% & 3.547 & -52.407\% & 3.208& 0.29(3.86\%) \\
		SVD-LoRA \cite{LoRA_comp}&CVPR'25
		&  -50.162\% & \phantom{-}7.920  &  -39.927\% & 2.207  &  -42.431\% & 1.938   & \underline{0.09(1.20\%)} \\

		\cmidrule(lr){1-9}
		\rowcolor{blue!4}Ours-Tiny&--
				& -83.187\% & 16.118 & -58.236\% & 3.742 & -55.387\% & 3.266& \textbf{0.08(1.07\%)} \\
		\rowcolor{blue!8}Ours-Small&--
				& \underline{-91.570\%} & \underline{16.934}  & \underline{-60.575\%} & \underline{3.980} & \underline{-60.661\%} & \underline{3.426}& 0.15(2.00\%) \\
		\rowcolor{blue!12}Ours-Base&--
				& \textbf{-92.788\%} & \textbf{17.575} & \textbf{-65.607\%} & \textbf{4.249} & \textbf{-62.589\%} & \textbf{3.624}& 0.32(4.26\%) \\
		\bottomrule
	\end{tabular}

	\label{tab:bdresults}
\end{table*}
\section{Experiments}

\subsection{Experimental Setup}

\subsubsection{Datasets.}
We evaluate the proposed method on three representative downstream machine vision tasks: image classification, object detection, and instance segmentation. We utilize the ImageNet dataset \cite{deng2009imagenet} for the classification, while the COCO2017 dataset \cite{lin2014microsoft} is used for object detection and instance segmentation.

\subsubsection{Benchmarks.}
To assess the effectiveness of the proposed method, we integrate CrossMambaTuning into two different base codecs. We use a simplified version of the Transformer-based Lu2022-TIC \cite{lu2021transformer} codec, which utilizes only the hyperprior entropy model. Additionally, we incorporate the CNN-based ELIC \cite{DBLP:conf/cvpr/HeYPMQW22} model with the Spatial-Channel Context model. To benchmark performance, we conduct comprehensive comparisons against recent SOTA methods, including Channel Selection \cite{liu2022improving}, ICMH-Net \cite{liu2023icmh}, TransTIC \cite{chen2023transtic}, Adapt-ICMH \cite{li2024image}, and SVD-LoRA \cite{LoRA_comp}. The entire framework is implemented based on CompressAI \cite{begaint2020compressai}.

\subsubsection{Evaluation Metrics.}
We measure compression performance using bits per pixel (bpp). To assess downstream task performance, we evaluate Top-1 accuracy on ImageNet-VAL (using a pre-trained ResNet-50 \cite{DBLP:conf/cvpr/HeZRS16}) and mAP on COCO2017-val. For object detection, we use Faster R-CNN \cite{DBLP:conf/nips/RenHGS15}, and for instance segmentation, we use Mask R-CNN \cite{he2017mask}, both implemented with Detectron2 \cite{wu2019detectron2}. Additionally, we report BD-rate \cite{bjontegaard2001calculation} and BD-mAP, which quantify bitrate savings at equivalent performance and performance gains at equivalent bitrates, respectively. 

\subsubsection{Training Details.}
In all experiments, the backbone remains frozen, and only the proposed adapters are optimized. All training images are randomly cropped to 256$\times$256. For classification, we train for 8 epochs with a batch size of 16. Conversely, for detection and segmentation tasks, we train for 40 epochs with a batch size of 8. Further details are available in Appendix H.
﻿

\subsection{Experimental Results}
\begin{table*}[!ht]
	\centering
	\caption{Ablation study of core components. 
	}
	\begin{tabular}{c ccccc cc cc c}
		\toprule
		\multirow{2}{*}{\textbf{Method}} 
		& \multicolumn{4}{c}{\textbf{Components}} 
		& \multicolumn{2}{c}{\textbf{Object Detection}} 
		& \multicolumn{2}{c}{\textbf{Instance Segmentation}} 
		& \multirow{2}{*}{\makecell{\textbf{Trainable}\\\textbf{Params}$\downarrow$ (M)}} \\
		\cmidrule(lr){2-5} \cmidrule(lr){6-7} \cmidrule(lr){8-9}
		&APC & LIE & SICA & TSPG & BD-Rate$\downarrow$ & BD-mAP$\uparrow$ & BD-Rate$\downarrow$ & BD-mAP$\uparrow$ & \\
		
		\midrule
		(a)& & & & & -60.162\% & 3.948 & -59.676\% & 3.405 & 0.26 (3.46\%) \\
		(b)&$\checkmark$ & & & & -61.303\% & 4.021 & -60.885\% & 3.479 & 0.26 (3.46\%) \\
		
		(c)&$\checkmark$ & $\checkmark$ & & & -61.977\% & 4.062 & -61.789\% & 3.493  & 0.27 (3.60\%) \\
		
		(d)&$\checkmark$ & $\checkmark$ & $\checkmark$ & & -64.931\% & 4.195 & -62.164\% & 3.575 & 0.30 (3.99\%) \\
		
		\rowcolor{blue!10}(e)&$\checkmark$ & $\checkmark$ & $\checkmark$ & $\checkmark$ & \textbf{-65.607\%} & \textbf{4.249} & \textbf{-62.589\%} & \textbf{3.624} & 0.32 (4.26\%) \\
		\bottomrule
	\end{tabular}
	\label{tab:ablation_core}
\end{table*}

\begin{table}[!ht]
	\centering

	\caption{Performance comparison of different methods on object detection, using ELIC \cite{DBLP:conf/cvpr/HeYPMQW22} as the base codec. 
	}
	\begin{tabular}{%
			c
			cc
			r
		}
		\toprule
		\multirow[c]{1}{*}{\textbf{Method}}
		& \multicolumn{2}{c}{\makecell{\textbf{Object Detection}\\BD-rate$\downarrow$ \enspace BD-mAP$\uparrow$}}
		& \multirow[t]{1}{*}{\makecell{\textbf{Trainable}\\\textbf{Params} $\downarrow$(M)}} 
		\\
		
		\midrule
		full fine-tuning
		& -70.409\% & 5.745 & 33.79(100.00\%) \\
		\cmidrule(lr){1-4}
		channel selection \cite{liu2022improving}
		& -23.235\%   & 1.518 & 2.68(7.93\%) \\
	
		SVD-LoRA \cite{LoRA_comp}
		& -41.190\% & 2.149 & 0.41(1.21\%) \\
		Adapt-ICMH \cite{li2024image}
		& -54.844\% & 3.379 & 0.41(1.21\%) \\
		\rowcolor{blue!4}Ours-Tiny 
		& -57.495\% & 3.863 & \textbf{0.11(0.33\%)} \\
		\rowcolor{blue!8}Ours-Small
		& \underline{-60.016\%} & \underline{4.137}  & \underline{0.21(0.62\%)} \\
		\rowcolor{blue!12}Ours-Base
		& \textbf{-62.710\%} & \textbf{4.500}  & 0.43(1.27\%) \\
		\bottomrule
	\end{tabular}

	\label{tab:elic_bd1}
\end{table}
Figure \ref{fig:main_tic} illustrates the rate-accuracy curves on the TIC codec, demonstrating that CrossMambaTuning consistently outperforms all competing methods. Notably, for classification, our "Small" model surpasses the state-of-the-art Adapt-ICMH while utilizing only $51.7\%$ of the trainable parameters. In detection and segmentation, even our "Tiny" variant significantly outperforms recent SOTA algorithms (Adapt-ICMH and SVD-LoRA). 

Quantitative results in Table \ref{tab:bdresults} corroborate this superiority: on detection and segmentation tasks, our "Tiny" variant outperforms Adapt-ICMH with merely $27\%$ of the parameters and exceeds SVD-LoRA by $1.328 \sim 1.533$ mAP at equivalent bitrates. Additional experiments on the CNN-based ELIC\cite{DBLP:conf/cvpr/HeYPMQW22} backbone further demonstrate this efficiency. As reported in Table~\ref{tab:elic_bd1}, the proposed method consistently surpasses existing state-of-the-art approaches. At the same bitrate, the "Tiny" variant improves mAP by $0.484$ for object detection, while reducing the number of trainable parameters by $73\%$. These results highlight both the parameter efficiency and generalization capability of CrossMambaTuning. 
Our "Tiny" and "Small" variants surpass larger SOTA methods in performance while utilizing fewer trainable parameters. This confirms both the high parameter efficiency and analytical soundness of our approach.

\begin{figure*}[!t]
  \centering
    \centerline{\includegraphics[width=0.9\linewidth]{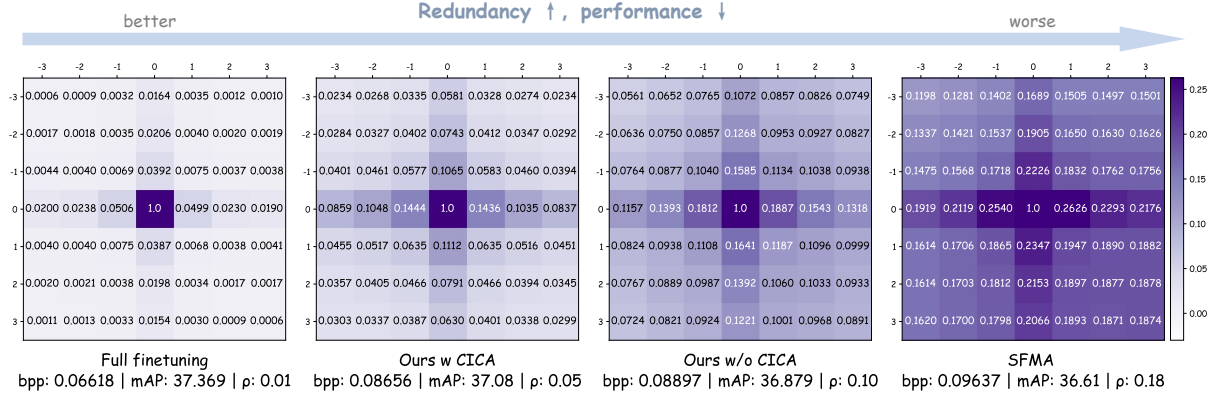}}
    \caption{
Visualization of spatial correlation for normalized latent representations $(y - \mu) / \sigma$. SFMA (right) exhibits significant spatial correlation. In contrast, our method (second from left) effectively suppresses correlations, approaching the decorrelation efficiency of Full Fine-tuning (left). More details are provided in Appendix E.
    }
    \label{heatmap_cor_map}
\end{figure*}
\subsection{Ablation Study}
\subsubsection{Core components.} 
As shown in Table \ref{tab:ablation_core}, we validate the effectiveness of our design through progressive integration. Compared to the baseline (Row a), the introduction of APC (Row b) achieves a significant improvement in detection metrics. Incorporating the LIE (Row c) then confirms the importance of multi-scale feature extraction. Subsequently, SICA (Row d) further reduces the BD-rate and demonstrates that cross-layer interaction reduces redundancy. Finally, the inclusion of the TSPG module (Row e) yields the best performance via task-specific prompt. These results demonstrate the effectiveness and complementarity of all components.
\subsubsection{Parameter sharing strategy.}
We evaluate the efficacy of the parameter sharing strategy within the SICA module on the detection task. As shown in Table \ref{tab:ablation_sharing}, disabling cross-layer sharing (instantiating independent modules for each bypass connection) increases the parameter count from 0.32M to 0.36M, yet leads to inferior performance. This suggests that, although different layers process features at varying semantic levels, they may still contain similar task-relevant information. The proposed sharing strategy therefore avoids repeatedly learning similar mappings across layers, reducing redundancy while preserving model compactness.
﻿

\begin{table}[!ht]
	\centering
	\caption{Ablation study on the parameter sharing strategy. 
	}
	\begin{tabular}{c ccc}
		\toprule
		\multirow{2}{*}{\textbf{Strategy}} 
		& \multicolumn{2}{c}{\textbf{Object Detection}} 
		& \multirow{2}{*}{\makecell{\textbf{Trainable}\\\textbf{Params}$\downarrow$ (M)}} \\
		\cmidrule(lr){2-3}
		& BD-Rate$\downarrow$ & BD-mAP$\uparrow$ & \\
		
		\midrule
		Independent & -63.224\% & 4.106 & 0.36 \\
		
		\rowcolor{blue!10}Shared (Ours) & \textbf{-65.607}\% & \textbf{4.249} & \textbf{0.32} \\
		\bottomrule
	\end{tabular}
	\label{tab:ablation_sharing}
\end{table}

\subsubsection{Local Perception Mechanism.} We investigate kernel configurations for local feature extraction. As shown in Table \ref{tab:ablation_local_perception}, while expanding to a  $5\times5$ kernel improves over the baseline, the dual-scale $DW1+DW3$ strategy outperforms the single large kernel. This proves that aggregating multi-grained features is more effective than simple spatial expansion. Ultimately, our multi-scale design achieves optimal performance by fusing diverse receptive fields.

\begin{table}[!ht]
	\centering
	\caption{Ablation study on the Local Perception Mechanism. 
	}
	\begin{tabular}{l ccc}
		\toprule
		\multirow{2}{*}{\textbf{\makecell{Kernel \\Configurations}}}
		& \multicolumn{2}{c}{\textbf{Object Detection}}
		& \multirow{2}{*}{\makecell{\textbf{Trainable}\\\textbf{Params}$\downarrow$ (M)}}\\
		\cmidrule(lr){2-3}
		& BD-Rate$\downarrow$&BD-mAP$\uparrow$ \\

		\midrule
		Baseline ($DW_{3}$) & -61.302\% & 4.021 & \textbf{0.26} \\
		
		Larger Kernel ($DW_{5}$) & -61.268\% & 4.032 & 0.26 \\
		
		Dual Scale ($DW_{1,3}$) & -61.678\% & 4.045 & 0.26 \\
		
		\rowcolor{blue!10}Ours ($DW_{1,3,5}$) & \textbf{-61.977}\% & \textbf{4.062} & 0.27 \\
		\bottomrule
	\end{tabular}
	\label{tab:ablation_local_perception}
\end{table}

\subsubsection{Spatial Modeling Module} To isolate the effect of the SSM module, we replace it with a window-based self-attention \cite{liu2021swin} module under the same experimental settings. As shown in Table~\ref{tab:ablation_spatial}, the window-attention variant performs slightly worse than the Mamba-based design, suggesting that Mamba is more effective for spatial modeling in machine-vision adaptation.

\begin{table}[!ht]
	\centering
	\caption{Ablation study on the spatial modeling module. 
	}
	\begin{tabular}{c ccc}
		\toprule
		\multirow{2}{*}{\textbf{Module}} 
		& \multicolumn{2}{c}{\textbf{Object Detection}} 
		& \multirow{2}{*}{\makecell{\textbf{Trainable}\\\textbf{Params}$\downarrow$ (M)}} \\
		\cmidrule(lr){2-3}
		& BD-Rate$\downarrow$ & BD-mAP$\uparrow$ & \\
		
		\midrule
		w/ window attention
				 & -62.020\% & 4.084 & 0.31 \\
		\rowcolor{blue!10}Ours (Mamba-based) & \textbf{-64.931}\% & \textbf{4.195} & \textbf{0.30} \\
		\bottomrule
	\end{tabular}
	\label{tab:ablation_spatial}
\end{table}

\subsection{Analysis of Redundancy Reduction.} We use spatial correlation as a proxy to quantify redundancy, following prior works that explicitly analyze redundancy through correlation maps and correlation values \cite{zhu2022transformer,DBLP:conf/nips/AliKQLKZBK23,DBLP:journals/corr/abs-2405-15413,DBLP:journals/corr/abs-2502-04988}, as well as recent designs that improve compression efficiency by reducing local correlation \cite{DBLP:conf/cvpr/0003CWWHLS25}. As shown in Fig.~\ref{heatmap_cor_map}, the previous SOTA method \cite{li2024image} exhibits relatively  high redundancy ($0.18$). In contrast, our approach reduces this metric to $0.05$, substantially narrowing the gap towards full fine-tuning ($0.01$). The increased correlation in the ablation variant (w/o SICA) further supports the role of the proposed SICA in alleviating redundancy.

\subsection{Reconstruction Quality Check}
Although our primary focus is downstream machine vision adaptation, we include a brief reconstruction check for completeness. Table~\ref{tab:recon_check} reports the bitrate and reconstruction quality of the base codec and the adapter-removed setting. Additional results for the adapter-attached setting are provided in Appendix B. Since our method features a plug-in design, removing the adapters exactly restores the pretrained codec, yielding identical RD performance.
\begin{table}[t]
\centering
\caption{Reconstruction quality on Imagenet-val. We report bitrate and reconstruction quality for different settings.
}
\label{tab:recon_check}
\begin{tabular}{lccc}
\toprule
\textbf{Method} & \textbf{bpp} & \textbf{PSNR(dB)} & \textbf{MS-SSIM} \\
\midrule
Base codec & 0.477 & 32.04 & 0.981 \\
Ours w/o adapters & 0.477 & 32.04 & 0.981 \\
\bottomrule
\end{tabular}
\end{table}

\subsection{Computational complexity and efficiency}
We benchmark the computational complexity of our method against full fine-tuning (full ft) in Table \ref{tab:complexity_analysis}. Our proposed variants (Ours-Base, Ours-Small, Ours-Tiny) demonstrate strong parameter efficiency: notably, the "Tiny" variant requires only $0.08$M trainable parameters during training, representing a substantial reduction compared to the $7.51$M required by full ft. 

Overall, our framework focuses on trainable parameter efficiency. It substantially reduces the trainable parameter count while remaining competitive in computational cost and latency. More details are provided in Appendix C.

\begin{table}[!ht]
	\centering
	\caption{We compare the complexity and performance of our method against the full fine-tuning baseline. The average encoding and decoding latencies are evaluated on an INTEL® XEON® SILVER 4310 CPU and an NVIDIA 4090 GPU.}
	\begin{tabular}{l cc cccc}
		\toprule
		\multirow{2}{*}{\textbf{Method}} 
		& \multicolumn{2}{c}{\textbf{KMACs/pixel}} 
		& \multicolumn{2}{c}{\textbf{Latency(ms)}} 
		& \multirow{2}{*}{\makecell{\textbf{Params}\\$\downarrow$(M)}} 
		& \multirow{2}{*}{\makecell{\textbf{BD}-\\\textbf{Acc}$\uparrow$}} \\
		\cmidrule(lr){2-3} \cmidrule(lr){4-5}
		& Enc & Dec & Enc & Dec & & \\
		
		\midrule
		full ft & 130.5 & 177.0 & 29.7 & 26.8 & 7.51 & 17.7 \\
		\rowcolor{blue!4}Ours-T & 137.9 & 185.9 & 32.2 & 29.1 & \textbf{0.08} & 16.1 \\
		\rowcolor{blue!8}Ours-S & 144.5 & 193.4 & 32.3 & 29.7 & \underline{0.15} & 16.9 \\
		\rowcolor{blue!12}Ours-B & 157.8 & 208.6 & 33.3 & 30.5 & 0.32 & 17.6 \\
		\bottomrule
	\end{tabular}
	\label{tab:complexity_analysis}
\end{table}

\begin{figure}[!htbp]
	\centering
	
	\includegraphics[width=0.95\linewidth]{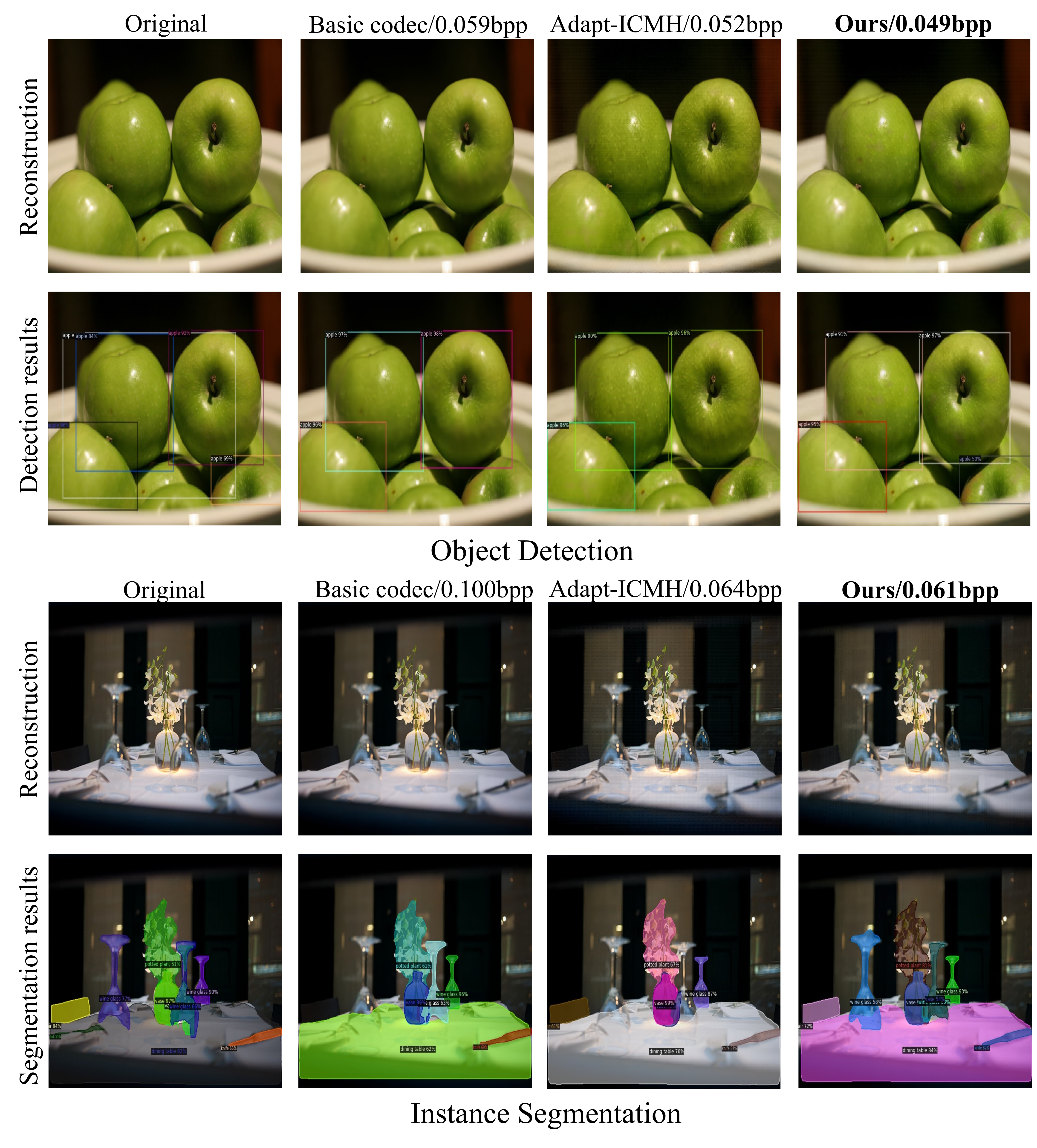}
	
	\caption{Qualitative comparison of different methods for detection and segmentation. The figure displays (from top to bottom): the original and decoded images, their corresponding results. 
	}

	\label{fig:visual_det_seg}
\end{figure}

\subsection{Qualitative Results}
Figure \ref{fig:visual_det_seg} presents a qualitative comparison demonstrating the superiority of our method. Compared to the SOTA method (Adapt-ICMH), our approach preserves more fine-grained details and task-relevant features, even at lower bitrates. Specifically, in the object detection task, our framework successfully identifies an apple instance that was missed by Adapt-ICMH. Similarly, for instance segmentation, it accurately segments a glass that the competing method failed to recognize. These results validate the effectiveness of the proposed CrossMambaTuning framework.

%
\section{Conclusion}
In this paper, we propose CrossMambaTuning, a novel parameter-efficient fine-tuning framework for adapting pre-trained image compression codecs to downstream machine vision tasks. Specifically, we introduce a Task-aware Mamba Adapter and a Scale-Invariant Cross-Layer Adapter to facilitate the efficient adaptation of pre-trained codecs. Extensive experimental results demonstrate that our method significantly outperforms existing approaches across multiple downstream tasks, including image classification, instance segmentation, and object detection. Notably, our lightweight variants (i.e., Small and Tiny) surpass current SOTA methods while requiring substantially fewer trainable parameters, thereby verifying both the parameter efficiency and effectiveness of the proposed framework. Furthermore, additional experiments across diverse architectures confirm the robustness of our approach.

\section{Acknowledgments}

The work was supported by the China Postdoctoral Science Foundation under Grant 2024M752531 and the National Natural Science Foundation of China under Grant 62171342.

\bibliographystyle{ACM-Reference-Format}
\balance
\bibliography{aaai2026}

\clearpage
\onecolumn
\setcounter{section}{0}
\renewcommand{\thesection}{\Alph{section}}
\section{Scope and Practical Relevance}
This work is motivated by practical multimedia deployment scenarios in which an on-device compression system is expected to support both human-vision image compression and machine vision compression under limited storage and deployment budgets. Instead of training and maintaining two separate learned compression systems from scratch, our framework reuses an existing pretrained LIC codec as the base system and performs parameter-efficient adaptation only for the machine vision mode. As a result, supporting machine vision compression requires training and storing only a small number of additional parameters, while the original codec remains available for human-vision compression.

\section{Reconstruction Quality Under Different Adapter Settings}
Table~\ref{tab:recon_check1} reports reconstruction reference results on ImageNet-val. The first two rows verify that removing the adapters exactly recovers the original pretrained codec, yielding identical bitrate and reconstruction quality. The remaining rows correspond to machine-vision compression methods, where the distortion term is optimized using task-oriented feature loss rather than the MSE-based reconstruction loss. Consequently, PSNR and MS-SSIM are not explicitly preserved during training, and reconstruction quality generally declines when the adapters remain attached.
﻿

For instance, compared to the base codec, the  adapter-attached setting decreases the PSNR from 28.92 dB to 23.90 dB and the MS-SSIM from 0.971 to 0.923. Comparable degradation is evident in other compression frameworks for machine vision, indicating that this is a prevalent consequence of optimizing for specific tasks instead of an anomaly specific to our approach. Furthermore, we note a qualitative tendency where methods exhibiting marginal improvements on machine tasks, like TransTIC\cite{chen2023transtic} and SVD-LoRA\cite{LoRA_comp}, maintain proportionally higher PSNR and MS-SSIM scores. This may indicate that their adaptations introduce weaker perturbations to the original representation, although we do not establish a strict causal relationship due to the bitrate and settings differences across methods.
﻿

In this PEFT-based setting, the adapter-based path is optimized for machine-vision compression, while the pretrained codec remains the path for reconstruction. Accordingly, deployment can be naturally interpreted as two selectable operating modes within the same LIC system: the pretrained codec path is used for human-vision compression, whereas the adapter-based path is used for machine-vision compression. Switching between the two modes only requires enabling or disabling a small set of task-specific parameters. Therefore, the results in Table~\ref{tab:recon_check1} should be interpreted as reconstruction references under machine-oriented optimization, rather than as part of a rate-distortion comparison.
﻿
﻿

\begin{table}[h]
\centering
\caption{Reconstruction reference for the ImageNet-val dataset. We report bitrate and reconstruction quality for the base codec, the adapter-removed setting, and several machine-vision compression methods. Since these methods are optimized using task-oriented feature loss rather than MSE loss, the reported PSNR and MS-SSIM values are included only as reconstruction references, not for a full rate-distortion comparison.}
\label{tab:recon_check1}
\begin{tabular}{lccc}
\toprule
\textbf{Method} & \textbf{bpp} & \textbf{PSNR(dB)} & \textbf{MS-SSIM} \\
\midrule
Base codec & 0.227 & 28.92 & 0.971 \\
Ours w/o adapters & 0.227 & 28.92 & 0.971 \\
\midrule
full ft & 0.239 & 22.66 & 0.903 \\
Ours w/ adapters & 0.206 & 23.90 & 0.923 \\
Adapt-ICMH \cite{li2024image}& 0.216 & 24.23 & 0.919 \\
TransTIC \cite{chen2023transtic}& 0.176 & 26.64 & 0.948 \\
SVD-LoRA \cite{LoRA_comp}& 0.190 & 28.31 & 0.953 \\
\bottomrule
\end{tabular}
\end{table}

\section{Additional Complexity Comparison with PEFT Baselines}
Table~\ref{tab:complexity_analysis_1} further compares CrossMambaTuning with full fine-tuning and representative PEFT baselines in terms of trainable parameters, computational cost, runtime latency, and downstream performance. All compared baselines follow their official implementations whenever available, and are evaluated under the same hardware environment as our method. The results indicate that the main advantage of our framework lies in parameter efficiency rather than in lower runtime cost. Relative to full fine-tuning, our variants substantially reduce the number of trainable parameters while preserving competitive performance, with moderate increases in KMACs and latency. Relative to existing PEFT baselines, CrossMambaTuning exhibits a more favorable parameter--performance trade-off: our ``Small" variant matches SFMA-64 in BD-Acc with substantially fewer trainable parameters, whereas our ``Tiny" variant outperforms SVD-LoRA by a large margin under a similarly small parameter budget. TransTIC, by comparison, incurs notably higher encoding cost and inferior machine-vision performance. Taken together, these results characterize CrossMambaTuning as a parameter-efficient machine-vision compression framework, rather than a method tailored for minimum inference complexity.
\begin{table}[!ht]
	\centering
\caption{Complexity and performance comparison with full fine-tuning and representative PEFT baselines. Average encoding and decoding latencies are measured on an INTEL® XEON® SILVER 4310 CPU and an NVIDIA 4090 GPU.}

	\begin{tabular}{l cc cccc}
		\toprule
		\multirow{2}{*}{\textbf{Method}} 
		& \multicolumn{2}{c}{\textbf{KMACs/pixel}} 
		& \multicolumn{2}{c}{\textbf{Latency(ms)}} 
		& \multirow{2}{*}{\makecell{\textbf{Params}\\$\downarrow$(M)}} 
		& \multirow{2}{*}{\makecell{\textbf{BD}-\\\textbf{Acc}$\uparrow$}} \\
		\cmidrule(lr){2-3} \cmidrule(lr){4-5}
		& Enc & Dec & Enc & Dec & & \\
		
		\midrule
		full ft & 130.5 & 177.0 & 29.7 & 26.8 & 7.51 & 17.7 \\
		Adapt-ICMH \cite{li2024image} & 145.2 & 191.7 & 31.4 & 28.5 & 0.29 & 16.9 \\
		TransTIC \cite{chen2023transtic}& 328.3 & 177.0 & 46.9 & 27.2 & 0.29 & 10.0 \\
		SVD-LoRA \cite{LoRA_comp}& 130.5 & 177.0 & 30.0 & 26.9 & \underline{0.09} & 7.9 \\
		\rowcolor{blue!4}Ours-T & 137.9 & 185.9 & 32.2 & 29.1 & \textbf{0.08} & 16.1 \\
		\rowcolor{blue!8}Ours-S & 144.5 & 193.4 & 32.3 & 29.7 & 0.15 & 16.9 \\
		\rowcolor{blue!12}Ours-B & 157.8 & 208.6 & 33.3 & 30.5 & 0.32 & 17.6 \\
		\bottomrule
	\end{tabular}
	\label{tab:complexity_analysis_1}
\end{table}

\section{Task-Specific Perceptual Loss}

During the training phase, we employed task-specific perceptual loss ($D_{task}$) to train for downstream tasks:
\begin{equation}
\mathcal{L} = \mathcal{R} + \lambda \cdot \mathcal{D}_{\text{task}}.
\end{equation}
To remain consistent with previous work, we followed the approach of \cite{chen2023transtic}, using pre-trained downstream task models (ResNet50 \cite{DBLP:conf/cvpr/HeZRS16}, Faster-RCNN\citep{DBLP:conf/nips/RenHGS15}, Mask-RCNN\citep{he2017mask}) to extract features from the original image $x$ and the reconstructed image $\hat{x}$, and computed the mean squared error (MSE) between the features. Specifically, we divided the downstream tasks into two groups: classification and instance segmentation/object detection.

\begin{figure}[!htbp]
	\centering
	\includegraphics[width=0.85\linewidth]{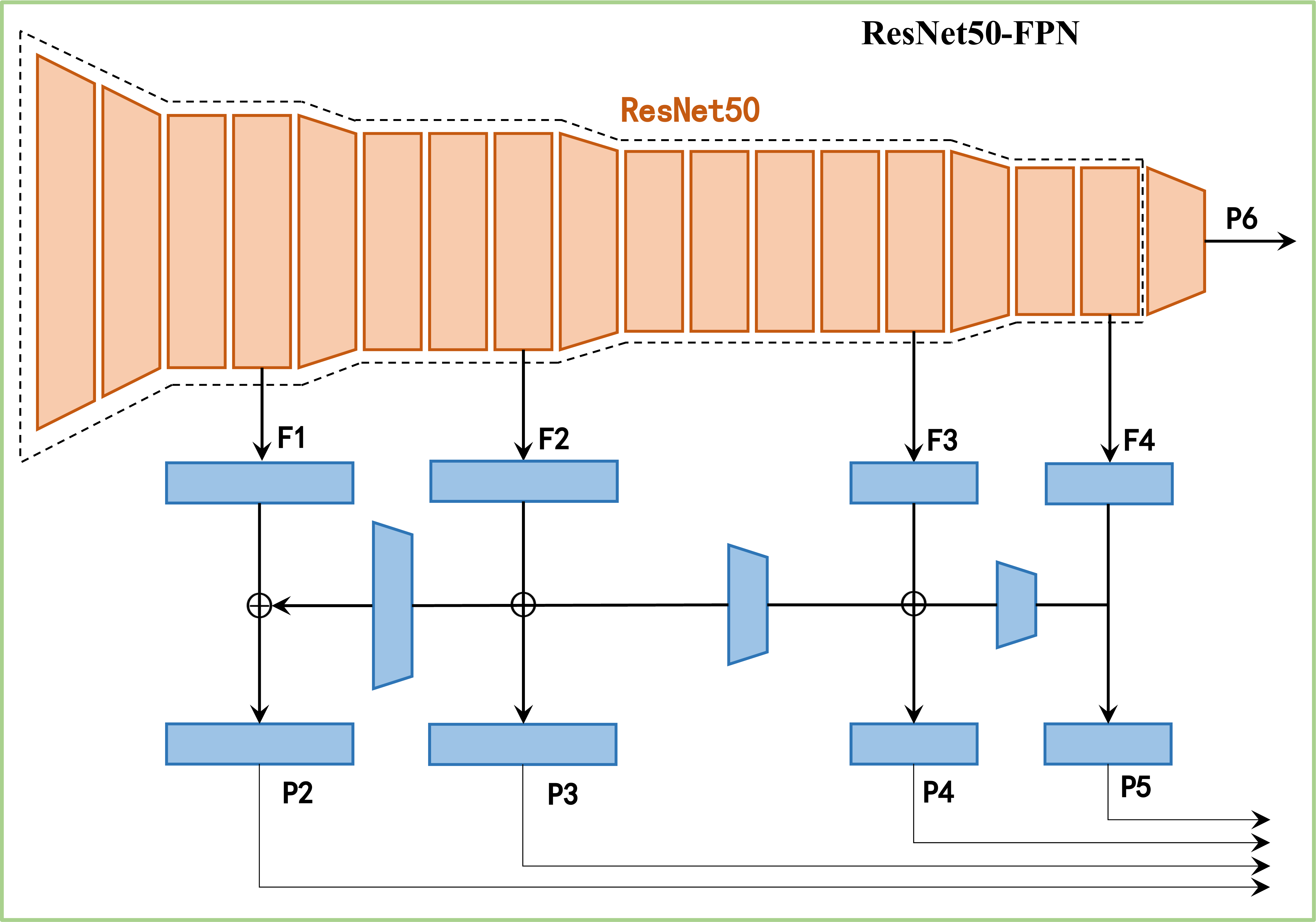}
	\hfill
	\caption{Network architecture of ResNet50-FPN.}
	\label{fig:sup_r50}
\end{figure}

We visually illustrate the feature layers used for loss computation in Figure \ref{fig:sup_r50}.

\subsection{Loss for Classification}
For the classification task, we utilize the feature layers from the ResNet-50 network \citep{DBLP:conf/cvpr/HeZRS16}. The perceptual loss is computed by averaging the MSE across feature layers $F_1, F_2, F_3, \text{and } F_4$. The loss formulation is defined as:

\begin{equation}
    D_{cls}(x, \hat{x}) = \frac{1}{4}\sum_{j=1}^{4}\text{MSE}(F_j(x), F_j(\hat{x})).
\end{equation}
For ResNet50, we used the V1 weights from Torchvision \cite{torchvision2016}, which achieved an accuracy of 76.7\% on the ImageNet-val dataset.
\subsection{Loss for Object Detection and Instance Segmentation}
For object detection and instance segmentation tasks, we employ Faster R-CNN \citep{DBLP:conf/nips/RenHGS15} and Mask R-CNN \citep{he2017mask}, respectively. Both architectures utilize the Feature Pyramid Network (FPN). Consequently, we compute the loss using feature levels $P_2, P_3, P_4, P_5, \text{and } P_6$. The corresponding loss function is calculated as:

\begin{equation}
    D_{det/seg}(x, \hat{x}) = \frac{1}{5}\sum_{j=2}^{6}\text{MSE}(P_j(x), P_j(\hat{x})).
\end{equation}

We used weights from Detectron2 \cite{wu2019detectron2} for both Faster R-CNN and Mask R-CNN. The mAP scores for detection and segmentation are 40.2\% and 37.2\%, respectively.

\section{Details of Spatial Correlation}

In Section 4, we  measured the spatial correlation $\rho$ of the proposed method and comparison method \cite{li2024image}. This section presents the formal definition of spatial correlation $\rho$ .

\subsection{Preliminaries and Model Settings}

Learned Image Compression (LIC) frameworks typically consist of the transform module and the entropy model. As stated in previous works \cite{balleVariationalImageCompression2018a,DBLP:journals/jstsp/BalleCMSJAHT21,zhu2022transformer}, an ideal transform module should effectively reduce the \textit{correlation} in the input original signal, enabling the use of simple scalar quantization methods and factorized entropy models without compromising performance.

Current LIC methods \cite{balleVariationalImageCompression2018a,minnenJointAutoregressiveHierarchical2018,zhu2022transformer,jiang2023mlicpp} predominantly adopt hyperprior-based entropy models (with most utilizing Gaussian entropy models). According to the aforementioned criterion, the transform module should aim to Gaussianize the distribution of the latent representation $y$ to minimize coding costs.

\begin{definition}
  \label{def:1}
Taking a simplified version of \citep{lu2021transformer} as an example, the latent representation $y$ is modeled as a Gaussian distribution $\mathcal{N}(\mu, \sigma)$, where the parameters are given by the hyperprior network $h_a$ and $h_s$. 
We define the normalized version of $y$ as $\overline{y}$:

\begin{equation}
    \overline{y} = \frac{y - \mu}{\sigma}.
\end{equation}
\end{definition}
In the ideal case, this results in a \textit{standard spherical normal vector}.

\subsection{Metric of Spatial Correlation}

\begin{definition}
\label{def:2}
For any two  variables $X$ and $Y$, the Pearson Correlation Coefficient is defined as the ratio of their covariance to the product of their standard deviations:
\begin{equation}
    \text{Pearson}(X, Y) = \frac{\text{Cov}(X, Y)}{\sigma_X \sigma_Y} = \frac{\mathbb{E}[(X - \mu_X)(Y - \mu_Y)]}{\sqrt{\mathbb{E}[(X - \mu_X)^2]} \sqrt{\mathbb{E}[(Y - \mu_Y)^2]}}.
\end{equation}
\end{definition}
This metric measures the degree of linear correlation between two variables and is invariant to scale.


\begin{definition}
Although $\overline{y}$  is theoretically designed to be a standard normal distribution, its actual distribution may exhibit deviations. To precisely evaluate the decorrelation effect, we use the Pearson correlation coefficient. Specifically, we compute the Pearson correlation coefficient for the normalized latent representation $\overline{y}$ at spatial positions $(w, h)$ and $(w+i, h+j)$, and then average it across the channel dimension to obtain the spatial correlation $\rho(i, j)$:

\begin{equation}
    \rho(i, j) = \frac{1}{C} \sum_{c=1}^{C} \text{Pearson}\left(\overline{y}_{w,h}^{(c)}, \overline{y}_{w+i, h+j}^{(c)}\right).
\end{equation}
\end{definition}

\begin{corollary} A lower value of spatial correlation $\rho$ indicates reduced redundancy across spatial positions in the latent representation, which is generally associated with improved rate--distortion performance. 
\end{corollary}

\textit{Analysis.}
Based on the above definitions, we employ spatial correlation as a metric to measure the redundancy within the  framework. Combined with the experimental results in Section 4, the proposed SICA module and Mamba-Adapter significantly reduced the $\rho$ value, thereby demonstrating that the method effectively reduces model redundancy.

\section{Metrics Definition}
\label{app:metrics}

In this work, we utilize the Wasserstein Distance and Kullback-Leibler (KL) Divergence to quantify the discrepancy between feature distributions. Let $P$ and $Q$ denote two probability distributions defined on the same metric space $\mathcal{X}$.

\subsection{Wasserstein Distance}
The Wasserstein distance, also known as the Earth Mover's Distance, measures the minimal cost of transporting probability mass between two distributions. The 1-Wasserstein distance is defined as
\begin{equation}
W_1(P, Q) = \inf_{\gamma \in \Pi(P, Q)} \mathbb{E}_{(x, y) \sim \gamma} [\|x - y\|] .
\end{equation}
Here, $\Pi(P, Q)$ denotes the set of all joint distributions whose marginals are $P$ and $Q$.
﻿

\subsection{Kullback-Leibler Divergence}
The Kullback-Leibler (KL) divergence measures the discrepancy between two probability distributions in terms of expected information content. It is defined as
\begin{equation}
    D_{KL}(P \| Q) = \int_{\mathcal{X}} p(x) \log \left( \frac{p(x)}{q(x)} \right) dx .
\end{equation}
The KL divergence characterizes the additional expected information cost incurred when distribution $Q$ is used to approximate distribution $P$. It therefore provides a quantitative measure of how much the approximation deviates from the reference distribution.

\section{Supplementary Information for CrossMambaTuning}

\begin{figure}[]
	\centering
	\includegraphics[width=1\linewidth]{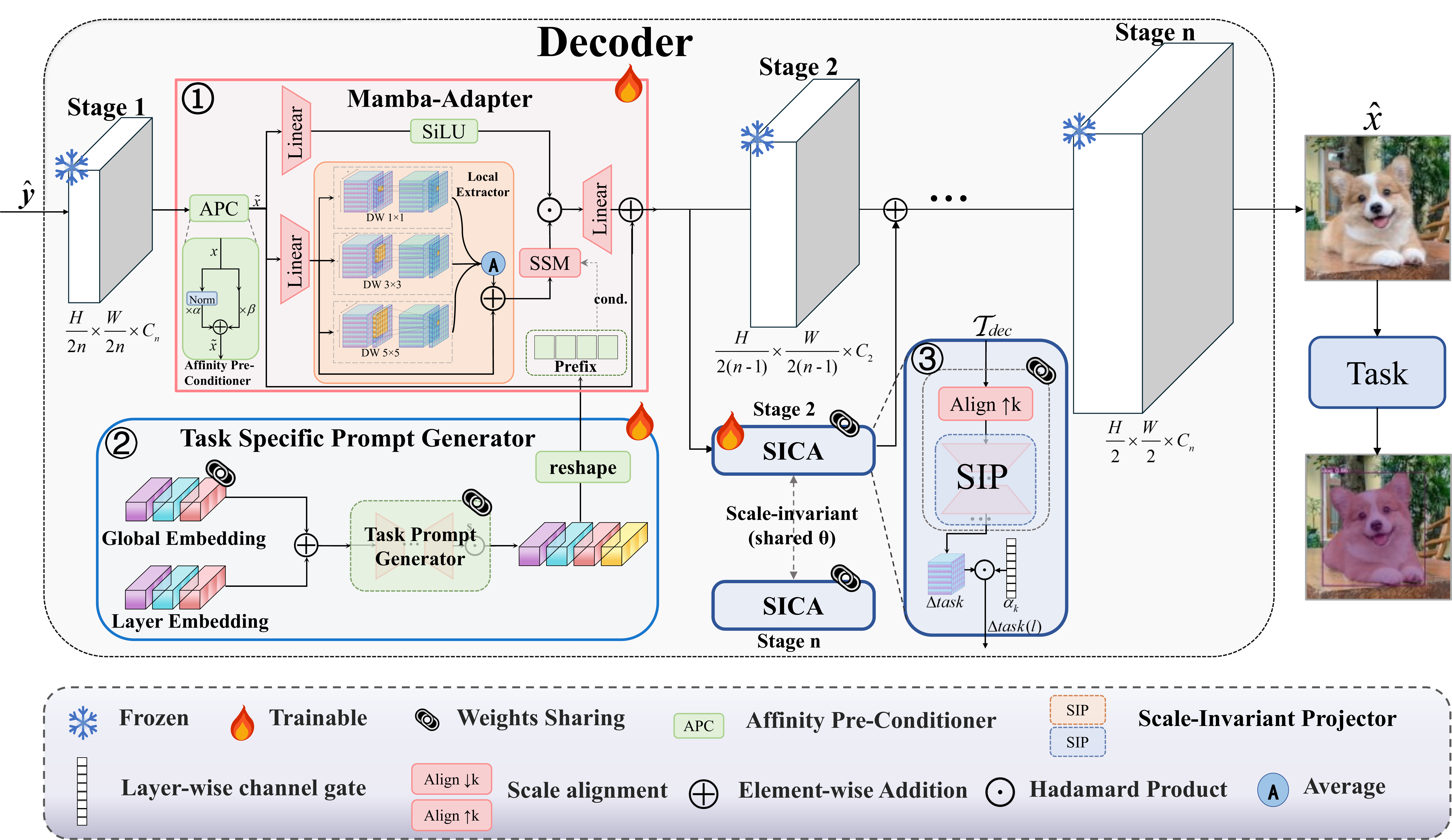}
	\hfill
	\caption{Overview of the proposed CrossMambaTuning (decoder stage). Efficient transfer is achieved by incorporating the proposed adapters into various pre-trained codecs. The snowflake symbol represents frozen layers, while the flame symbol represents trainable layers. }
	\label{fig:decoder}
\end{figure}

In this section, we illustrate the detailed architecture of CrossMambaTuning in the decoder stage. As depicted in Figure \ref{fig:decoder}, this operates as the inverse process of the encoder stage.
\section{More Implementation Details}
\subsection{Module Configuration}
In this section, we detail the specific configurations of the proposed CrossMambaTuning.

\subsubsection*{1. Base Configuration}
To achieve optimal performance, we integrate the APC, LIE, SICA, and TSPG modules into the Base model, constructing a comprehensive Task-Aware Mamba-Adapter and Cross-Layer Interaction framework. The specific configurations for each module are as follows:

\begin{itemize}
    \item The learnable parameters $\alpha$ and $\beta$ in the APC module are initialized to $1 \times 10^{-6}$ and $1$, respectively, to ensure stability during the initial training phase.
    \item The parameter $\alpha_k$ in the SICA module is initialized to $0$ to prevent interference with the optimization process of the Mamba-Adapter in the early stages of training.
    \item In the TSPG module, the task embeddings $E_{\text{task}}$ and layer embeddings $E_{\text{layer}}$ are initialized using a Gaussian distribution with a mean of $0$ and a variance of $0.02$. The intermediate dimension of its bottleneck MLP is set to $32$.
    \item The intermediate dimension of the Base model is set to $64$. Specifically, input features are first projected down to $64$ after passing through the APC module, then modulated in the spatial domain, and finally restored to their original dimensions.
    \item The state space dimension $d_{\text{state}}$ is set to $16$ to strike a favorable balance between computational cost and modeling capability.
\end{itemize}

We adopt SS2D \cite{DBLP:conf/nips/LiuTZYX0YJ024} as the selective scanning implementation in the Mamba Adapter, without introducing additional modifications to the scanning process. Our contribution is a Task-aware Mamba Adapter for machine vision compression, which introduces APC, LIE, and TSPG to improve distribution alignment, local dependency modeling, and task-aware conditioning, respectively.

Furthermore, to align with the information bottleneck design of the Mamba-Adapter, the intermediate dimension of the SICA module is set to $64$, thereby facilitating effective cross-layer feature adaptation.

\subsubsection*{2. Lightweight Configurations}
To construct more lightweight and training-stable models, we reduce the intermediate dimensions of the Mamba-Adapter and SICA modules to $32$ and $16$ for the Small and Tiny variants, respectively. Concurrently, we omit the TSPG module in these lightweight settings, as its additional gain becomes limited under stricter parameter budgets. As shown in Table~\ref{tab:ablation_tspg}, enabling TSPG in the Tiny variant brings only marginal additional improvement, while increasing the trainable parameters from $0.08$M to $0.09$M. Owing to these designs, the number of trainable parameters for the Small and Tiny models is reduced to $47\%$ and $25\%$ of the Base model, respectively, significantly improving parameter efficiency.

\begin{table}[!ht]
	\centering
	\caption{Ablation study on the TSPG in Tiny variant. 
	}
	\begin{tabular}{c ccc}
		\toprule
		\multirow{2}{*}{\textbf{Strategy}} 
		& \multicolumn{2}{c}{\textbf{Object Detection}} 
		& \multirow{2}{*}{\makecell{\textbf{Trainable}\\\textbf{Params}$\downarrow$ (M)}} \\
		\cmidrule(lr){2-3}
		& BD-Rate$\downarrow$ & BD-mAP$\uparrow$ & \\
		
		\midrule
		Ours-T & -58.236\% & 3.742 & \textbf{0.08} \\
		
		\rowcolor{blue!10}Ours-T w/ TSPG & \textbf{-58.453}\% & \textbf{3.749} & {0.09} \\
		\bottomrule
	\end{tabular}
	\label{tab:ablation_tspg}
\end{table}
\subsection{Settings and Hyperparameters}

All experiments were conducted using the PyTorch framework. The software environment and hardware configuration are summarized below.

\textbf{Software Setup.}
Experiments were performed on a Linux system running Ubuntu 20.04. PyTorch 2.4 was used as the deep learning framework, with CUDA 12.4 and cuDNN 8.9 providing GPU acceleration.

\textbf{Hardware Configuration.}
All models were trained on a single NVIDIA GeForce RTX 4090 GPU with 24~GB of VRAM. The training server was equipped with an Intel Xeon Silver 4310 CPU operating at 2.10~GHz and 256~GB of system memory.

The hyperparameter settings for different downstream tasks are reported in Table~\ref{tab:hyperparameters}. For all tasks, we employed the Adam optimizer with a base learning rate of $4\times10^{-4}$. Owing to task-specific characteristics, different batch sizes and training schedules were adopted for classification, object detection, and instance segmentation. Specifically, classification models were trained for 8 epochs, whereas detection and segmentation models were trained for a larger number of epochs using a fixed learning rate. The trade-off coefficient $\lambda$ was selected from task-specific candidate sets to balance rate and distortion objectives.

\begin{table}[h!]
	\centering
\caption{Training hyperparamters for experiments.}
	\begin{tabular}{l|ccc}
		\hline
		& Classification      & Detection         & Segmentation      \\ \hline
		Optimizer                  & Adam                & Adam              & Adam              \\
		Batch size                 & 16                  & 8                 & 8                 \\
		Trade-off term $\lambda$   & \makecell{{[}2.5, 3.5,\\ 5, 6.7, 13{]} }& \makecell{{[}0.5, 0.875, \\1.75, 3{]}} & \makecell{ {[}0.35, 0.5, \\0.875, 1.75, 3{]}} \\
		Epochs                     & 8                   & 40                & 40                \\
		Learning rate schedule     & -         & -                 & -                 \\
		Milestones                 & -             & -                 & -                 \\
		Learning rate decay        & -                & -                 & -                 \\
		Base learning rate         & 4e-4                & 4e-4              & 4e-4              \\ \hline
	\end{tabular}
	
	\label{tab:hyperparameters}
\end{table}

\section{Limitations}
Several boundary conditions of the current study are worth noting. First, our experiments focus on high-level downstream machine vision tasks, and the applicability of the framework to low-level vision tasks remains outside the present scope. Second, although CrossMambaTuning is parameter-efficient, its runtime cost still depends on the underlying codec and may require further simplification in extremely high-resolution or strict real-time settings. Third, our study does not focus on extremely low bitrate regimes (e.g., below 0.05 bpp), where the compressed representation itself may become the dominant bottleneck and limit the benefit of downstream adaptation. Finally, while the framework is validated on both CNN-based and Transformer-based codecs, broader evaluation on more diverse compression architectures and practical deployment environments remains future work.
﻿

\end{document}